\documentclass[11pt,a4paper]{article}
\usepackage[hyperref]{emnlp2018}
\usepackage{times}
\usepackage{latexsym}
\usepackage{amsmath, amssymb, amsthm}
\usepackage{booktabs}
\usepackage{graphicx}

\usepackage{xcolor}

\usepackage{pgfplots}
\pgfplotsset{width=0.47\textwidth,compat=1.9, height=2in}

\usepackage{url}
\usepackage[normalem]{ulem}
\usepackage{tikz-dependency, rotating}

\newcommand{\fone}{$\T{F}_1$}

\def\model#1{#1}

\def\f#1{\textrm{#1}}
\def\fb#1{\textbf{#1}}
\def\fu#1{\underline{#1}}

\newcommand\bh{\ensuremath{\mathbf{h}}}

\newcommand\bA{\ensuremath{\mathbf{A}}}

\newcommand\bI{\ensuremath{\mathbf{I}}}

\newcommand\BR{\ensuremath{\mathbb{R}}}

\newcommand\T{\text}

\newcommand\refeqn[1]{(\ref{eqn:#1})}

\newcommand\refsec[1]{Section~\ref{sec:#1}}

\newcommand\reffig[1]{Figure~\ref{fig:#1}}

\newcommand\reftab[1]{Table~\ref{tab:#1}}

\usepackage{color}

\def\rel#1{\text{\textit{#1}}}

\def\ner#1{\text{\texttt{#1}}}

\aclfinalcopy 

\definecolor{red}{HTML}{E31A1C}
\definecolor{blue}{HTML}{1F78B4}
\definecolor{green}{HTML}{33A02C}
\definecolor{orange}{HTML}{FF7F00}
\definecolor{purple}{HTML}{6A3D9A}

\newcommand{\red}[1]{{#1}}

\title{Graph Convolution over Pruned Dependency Trees\\ Improves Relation Extraction}

\author{Yuhao Zhang,*
Peng Qi,*
Christopher D. Manning
\\
Stanford University\\
Stanford, CA 94305\\
  {\tt \{yuhaozhang, pengqi, manning\}@stanford.edu}\\
\\}

\date{}

\begin{document}
\maketitle

\renewcommand{\thefootnote}{\fnsymbol{footnote}}
\footnotetext[1]{Equal contribution. The order of authorship was decided by a tossed coin.}
\renewcommand{\thefootnote}{\arabic{footnote}}

\setlength{\abovedisplayskip}{5pt}
\setlength{\belowdisplayskip}{5pt}
\setlength{\abovedisplayshortskip}{0pt}
\setlength{\belowdisplayshortskip}{0pt}

\begin{abstract}

Dependency trees help relation extraction models capture long-range relations between words.
However, existing dependency-based models either neglect crucial information (e.g., negation) by pruning the dependency trees too aggressively,
or are computationally inefficient because it is difficult to parallelize over different tree structures.
We propose an extension of graph convolutional networks that is tailored for relation extraction, which pools information over arbitrary dependency structures efficiently in parallel.
To incorporate relevant information while maximally removing irrelevant content, we further apply a novel pruning strategy to the input trees by keeping words immediately around the shortest path between the two entities among which a relation might hold.
The resulting model achieves state-of-the-art performance on the large-scale TACRED dataset, outperforming existing sequence and dependency-based neural models.
We also show through detailed analysis that this model has complementary strengths to sequence models, and combining them further improves the state of the art.

\end{abstract}

\section{Introduction}

Relation extraction involves discerning whether a relation exists between two entities in a sentence (often termed \emph{subject} and \emph{object}, respectively).
Successful relation extraction is the cornerstone of applications requiring relational understanding of unstructured text on a large scale, such as question answering \cite{yu2017improved},  knowledge base population~\cite{zhang2017tacred}, and biomedical knowledge discovery~\cite{quirk2016distant}.

\begin{figure}
	\centering
	\includegraphics[width=0.43\textwidth]{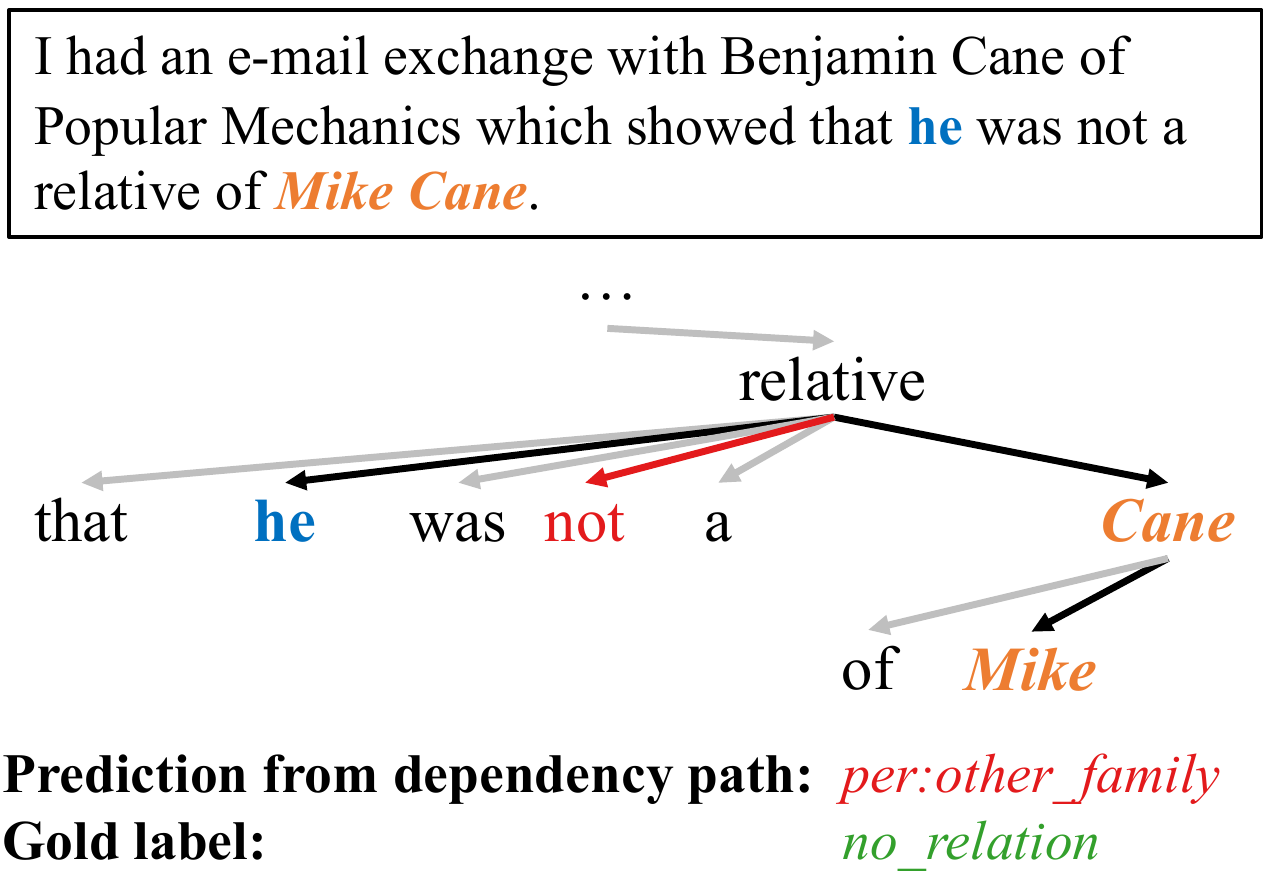}
	\caption{An example modified from the TAC KBP challenge corpus.
	A subtree of the original UD dependency tree between the subject (``he'') and object (``Mike Cane'') is also shown, where the shortest dependency path between the entities is highlighted in bold. 
	Note that negation (``not'') is off the dependency path.}
	\label{fig:example}
\end{figure}

Models making use of dependency parses of the input sentences, or \emph{dependency-based models}, have proven to be very effective in relation extraction, because they capture long-range syntactic relations that are obscure from the surface form alone (e.g., when long clauses or complex scoping are present).
Traditional feature-based models are able to represent dependency information by featurizing dependency trees as overlapping paths along the trees~\cite{kambhatla2004combining}.
However, these models face the challenge of sparse feature spaces and are brittle to lexical variations.
More recent neural models address this problem with distributed representations built from their computation graphs formed along parse trees.
One common approach to leverage dependency information is to perform bottom-up or top-down computation along the parse tree or the subtree below the lowest common ancestor (LCA) of the entities~\cite{miwa2016end}.
Another popular approach, inspired by \citet{bunescu2005shortest}, is to reduce the parse tree to the \emph{shortest dependency path} between the entities~\cite{xu2015semantic,xu2015classifying}.

However, these models suffer from several drawbacks.
Neural models operating directly on parse trees are usually difficult to parallelize and thus computationally inefficient, because aligning trees for efficient batch training is usually non-trivial.
Models based on the shortest dependency path between the subject and object are computationally more efficient, but this simplifying assumption has major limitations as well.
\reffig{example} shows a real-world example where crucial information (i.e., negation) would be excluded when the model is restricted to only considering the dependency path.

In this work, we propose a novel extension of the graph convolutional network~\cite{kipf2016semi, marcheggiani2017encoding} that is tailored for relation extraction.
Our model encodes the dependency structure over the input sentence with efficient graph convolution operations, then extracts entity-centric representations to make robust relation predictions.
We also apply a novel \emph{path-centric pruning} technique to remove irrelevant information from the tree while maximally keeping relevant content, which further improves the performance of several dependency-based models including ours.

We test our model on the popular SemEval 2010 Task 8 dataset and the more recent, larger TACRED dataset.
On both datasets, our model not only outperforms existing dependency-based neural models by a significant margin when combined with the new pruning technique, but also achieves a 10--100x speedup over existing tree-based models.
On TACRED, our model further achieves the state-of-the-art performance, surpassing a competitive neural sequence model baseline.
This model also exhibits complementary strengths to sequence models on TACRED, and combining these two model types through simple prediction interpolation further improves the state of the art.

To recap, our main contributions are:
(i) we propose a neural model for relation extraction based on graph convolutional networks, which allows it to efficiently pool information over arbitrary dependency structures;
(ii) we present a new path-centric pruning technique to help dependency-based models maximally remove irrelevant information without damaging crucial content to improve their robustness;
(iii) we present detailed analysis on the model and the pruning technique, and show that dependency-based models have complementary strengths with sequence models.

\section{Models}

In this section, we first describe graph convolutional networks (GCNs) over dependency tree structures, and then we introduce an architecture that uses GCNs at its core for relation extraction.

\subsection{Graph Convolutional Networks over Dependency Trees} \label{sec:gcn_for_dep}

The graph convolutional network \cite{kipf2016semi} is an adaptation of the convolutional neural network \citep{lecun1998gradient} for encoding graphs.
Given a graph with $n$ nodes, we can represent the graph structure with an $n \times n$ adjacency matrix $\bA$ where $A_{ij} = 1$ if there is an edge going from node $i$ to node $j$.
In an $L$-layer GCN, if we denote by $h_i^{(l-1)}$ the input vector and $h_i^{(l)}$ the output vector of node $i$ at the $l$-th layer, a graph convolution operation can be written as
\begin{align}
	h_i^{(l)} = \sigma\big( \sum_{j=1}^n A_{ij} W^{(l)}{h}_j^{(l-1)} + b^{(l)} \big),
	\label{eqn:conv}
\end{align}
where $W^{(l)}$ is a linear transformation, $b^{(l)}$ a bias term, and $\sigma$ a nonlinear function (e.g., ReLU).
Intuitively, during each graph convolution, each node gathers and summarizes information from its neighboring nodes in the graph.

We adapt the graph convolution operation to model dependency trees by converting each tree into its corresponding adjacency matrix $\bA$, where $A_{ij} = 1$ if there is a dependency edge between tokens $i$ and $j$.
However, naively applying the graph convolution operation in Equation~\refeqn{conv} could lead to node representations with drastically different magnitudes, since the degree of a token varies a lot.
This could bias our sentence representation towards favoring high-degree nodes regardless of the information carried in the node (see details in \refsec{relation}).
Furthermore, the information in $h^{(l-1)}_i$ is never carried over to $h^{(l)}_i$, since nodes never connect to themselves in a dependency tree.

We resolve these issues by normalizing the activations in the graph convolution before feeding it through the nonlinearity, and adding self-loops to each node in the graph:
\begin{align}
	h_i^{(l)} =& \sigma\big( \sum_{j=1}^n \tilde{A}_{ij} W^{(l)}{h}_j^{(l-1)} / d_i + b^{(l)} \big),\label{eqn:gcn_final}
\end{align}
where $\tilde{\bA}=\bA + \bI$ with $\bI$ being the $n\times n$ identity matrix, and $d_i=\sum_{j=1}^n \tilde{A}_{ij}$ is the degree of token $i$ in the resulting graph.

\begin{figure*}
	\centering
	\includegraphics[width=0.98\textwidth]{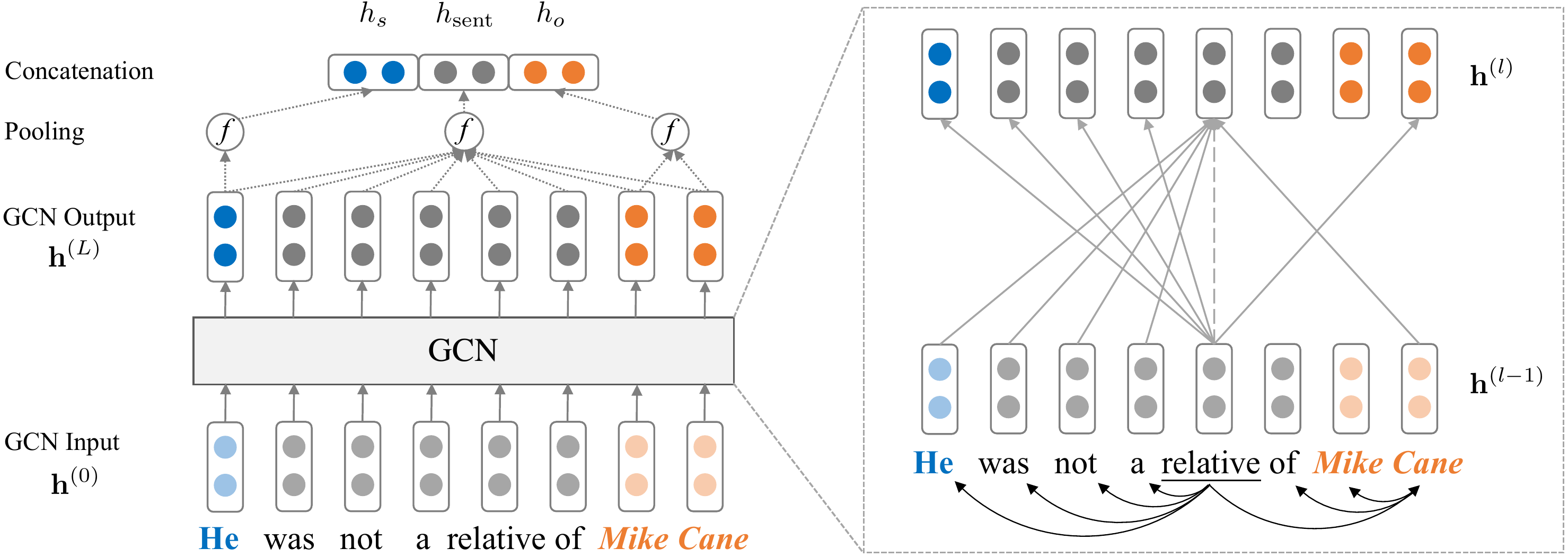}
	\caption{Relation extraction with a graph convolutional network. The left side shows the overall architecture, while on the right side, we only show the detailed graph convolution computation for the word ``relative'' for clarity. A full unlabeled dependency parse of the sentence is also provided for reference.}
	\label{fig:gcn}
\end{figure*}

Stacking this operation over $L$ layers gives us a deep GCN network, where we set $h_1^{(0)}, \ldots, h_n^{(0)}$ to be input word vectors, and use $h_1^{(L)}, \ldots, h_n^{(L)}$ as output word representations.
All operations in this network can be efficiently implemented with matrix multiplications, making it ideal for batching computation over examples and running on GPUs.
Moreover, the propagation of information between tokens occurs in parallel, and the runtime does not depend on the depth of the dependency tree.

Note that the GCN model presented above uses the same parameters for all edges in the dependency graph.
We also experimented with:
(1) using different transformation matrices $W$ for top-down, bottom-up, and self-loop edges;
and (2) adding dependency relation-specific parameters for edge-wise gating, similar to \cite{marcheggiani2017encoding}.
We found that modeling directions does not lead to improvement,\footnote{We therefore treat the dependency graph as undirected, i.e. $\forall i, j, A_{ij}=A_{ji}$.} and adding edge-wise gating further hurts performance.
We hypothesize that this is because the presented GCN model is usually already able to capture dependency edge patterns that are informative for classifying relations, and modeling edge directions and types does not offer additional discriminative power to the network before it leads to overfitting.
For example, the relations entailed by ``\emph{A}'s son, \emph{B}'' and ``\emph{B}'s son, \emph{A}'' can be readily distinguished with ``'s'' attached to different entities, even when edge directionality is not considered.

\subsection{Encoding Relations with GCN}
\label{sec:relation}
We now formally define the task of relation extraction.
Let $\mathcal{X} = [x_1, ..., x_n]$ denote a sentence, where $x_i$ is the $i^{\text{th}}$ token.
A subject entity and an object entity are identified and correspond to two spans in the sentence: $\mathcal{X}_s = [x_{s_1}, \ldots, x_{s_2}]$ and $\mathcal{X}_o = [x_{o_1}, \ldots, x_{o_2}]$.
Given $\mathcal{X}$, $\mathcal{X}_s$, and $\mathcal{X}_o$, the goal of relation extraction is to predict a relation $r \in \mathcal{R}$ (a predefined relation set) that holds between the entities or ``no relation'' otherwise.

After applying an $L$-layer GCN over word vectors, we obtain hidden representations of each token that are directly influenced by its neighbors no more than $L$ edges apart in the dependency tree.
To make use of these word representations for relation extraction, we first obtain a sentence representation as follows (see also \reffig{gcn} left):
\begin{align}
h_{\text{sent}} = f \big( \bh^{(L)} \big) = f\big( \text{GCN}(\bh^{(0)}) \big),
\label{eqn:hsent}
\end{align}
where $\bh^{(l)}$ denotes the collective hidden representations at layer $l$ of the GCN, and $f: \BR^{d \times n} \to \BR^{d}$ is a max pooling function that maps from $n$ output vectors to the sentence vector.

We also observe that information close to entity tokens in the dependency tree is often central to relation classification.
Therefore, we also obtain a subject representation $h_s$ from $\bh^{(L)}$ as follows
\begin{align}
	h_{s} = f \big( \bh_{s_1:s_2}^{(L)} \big),
\end{align}
as well as an object representation $h_o$ similarly.

Inspired by recent work on relational learning between entities~\cite{santoro2017simple,lee2017end}, we obtain the final representation used for classification by concatenating the sentence and the entity representations, and feeding them through a feed-forward neural network (FFNN):
\begin{align}
	h_{\text{final}} = \text{FFNN} \big( [h_{\text{sent}}; h_s; h_o] \big).
\end{align}
This $h_{\text{final}}$ representation is then fed into a linear layer followed by a softmax operation to obtain a probability distribution over relations.

\subsection{Contextualized GCN}

The network architecture introduced so far learns effective representations for relation extraction, but it also leaves a few issues inadequately addressed.
First, the input word vectors do not contain contextual information about word order or disambiguation.
Second, the GCN highly depends on a correct parse tree to extract crucial information from the sentence (especially when pruning is performed), while existing parsing algorithms produce imperfect trees in many cases.

To resolve these issues, we further apply a Contextualized GCN (C-GCN) model, where the input word vectors are first fed into a bi-directional long short-term memory (LSTM) network to generate contextualized representations, which are then used as $\bh^{(0)}$ in the original model.
This BiLSTM contextualization layer is trained jointly with the rest of the network.
We show empirically in \refsec{expr} that this augmentation substantially improves the performance over the original model.

We note that this relation extraction model is conceptually similar to graph kernel-based models \cite{zelenko2003kernel}, in that it aims to utilize local dependency tree patterns to inform relation classification.
Our model also incorporates crucial off-path information, which greatly improves its robustness compared to shortest dependency path-based approaches.
Compared to tree-structured models (e.g., Tree-LSTM \cite{tai2015improved}), it not only is able to capture more global information through the use of pooling functions, but also achieves substantial speedup by not requiring recursive operations that are difficult to parallelize.
For example, we observe that on a Titan Xp GPU, training a Tree-LSTM model over a minibatch of 50 examples takes 6.54 seconds on average, while training the original GCN model takes only 0.07 seconds, and the C-GCN model 0.08 seconds.

\section{Incorporating Off-path Information with Path-centric Pruning}
\label{sec:prune}

Dependency trees provide rich structures that one can exploit in relation extraction, but most of the information pertinent to relations is usually contained within the subtree rooted at the lowest common ancestor (LCA) of the two entities.
Previous studies \cite{xu2015classifying, miwa2016end} have shown that removing tokens outside this scope helps relation extraction by eliminating irrelevant information from the sentence.
It is therefore desirable to combine our GCN models with tree pruning strategies to further improve performance.
However, pruning too aggressively (e.g., keeping only the dependency path) could lead to loss of crucial information and conversely hurt robustness.
For instance, the negation in \reffig{example} is neglected when a model is restricted to only looking at the dependency path between the entities.
Similarly, in the sentence ``\emph{She was diagnosed with cancer last year, and succumbed this June}'', the dependency path \emph{She}$\gets$\emph{diagnosed}$\to$\emph{cancer} is not sufficient to establish that \emph{cancer} is the cause of death for the subject unless the conjunction dependency to \emph{succumbed} is also present.

Motivated by these observations, we propose \emph{path-centric pruning}, a novel technique to incorporate information off the dependency path.
This is achieved by including tokens that are up to distance $K$ away from the dependency path in the LCA subtree.
$K=0$, corresponds to pruning the tree down to the path,
$K=1$ keeps all nodes that are directly attached to the path,
and $K=\infty$ retains the entire LCA subtree.
We combine this pruning strategy with our GCN model, by directly feeding the pruned trees into the graph convolutional layers.%
\footnote{For our C-GCN model, the LSTM layer still operates on the full sentence regardless of the pruning.}
We show that pruning with $K=1$ achieves the best balance between including relevant information (e.g., negation and conjunction) and keeping irrelevant content out of the resulting pruned tree as much as possible.

\section{Related Work}

At the core of fully-supervised and distantly-supervised relation extraction approaches are statistical classifiers, many of which find syntactic information beneficial.
For example, \citet{mintz2009distant} explored adding syntactic features to a statistical classifier and found them to be useful when sentences are long.
Various kernel-based approaches also leverage syntactic information to measure similarity between training and test examples to predict the relation, finding that tree-based kernels \cite{zelenko2003kernel} and dependency path-based kernels \cite{bunescu2005shortest} are effective for this task.

Recent studies have found neural models effective in relation extraction.
\citet{zeng2014relation} first applied a one-dimensional convolutional neural network (CNN) with manual features to encode relations.
\citet{vu2016combining} showed that combining a CNN with a recurrent neural network (RNN) through a voting scheme can further improve performance.
\citet{zhou2016attention} and \citet{wang2016relation} proposed to use attention mechanisms over RNN and CNN architectures for this task.

Apart from neural models over word sequences, incorporating dependency trees into neural models has also been shown to improve relation extraction performance by capturing long-distance relations.
\citet{xu2015classifying} generalized the idea of dependency path kernels by applying a LSTM network over the shortest dependency path between entities.
\citet{liu2015dependency} first applied a recursive network over the subtrees rooted at the words on the dependency path and then applied a CNN over the path.
\citet{miwa2016end} applied a Tree-LSTM~\cite{tai2015improved}, a generalized form of LSTM over dependency trees, in a joint entity and relation extraction setting.
They found it to be most effective when applied to the subtree rooted at the LCA of the two entities.

More recently, \citet{adel2016comparing} and \citet{zhang2017tacred} have shown that relatively simple neural models (CNN and augmented LSTM, respectively) can achieve comparable or superior performance to dependency-based models when trained on larger datasets.
In this paper, we study dependency-based models in depth and show that with a properly designed architecture, they can outperform and have complementary advantages to sequence models, even in a large-scale setting.

Finally, we note that a technique similar to path-centric pruning has been applied to reduce the space of possible arguments in semantic role labeling \cite{he2018syntax}.
The authors showed pruning words too far away from the path between the predicate and the root to be beneficial, but reported the best pruning distance to be 10, which almost always retains the entire tree.
Our method differs in that it is applied to the shortest dependency path between entities, and we show that in our technique the best pruning distance is 1 for several dependency-based relation extraction models.

\section{Experiments}
\label{sec:expr}

\subsection{Baseline Models}

We compare our models with several competitive dependency-based and neural sequence models.
\vspace{-.1em}
\paragraph*{Dependency-based models.}
In our main experiments we compare with three types of dependency-based models.
(1) A logistic regression (LR) classifier which combines dependency-based features with other lexical features.
(2) Shortest Dependency Path LSTM (SDP-LSTM) \cite{xu2015classifying}, which applies a neural sequence model on the shortest path between the subject and object entities in the dependency tree.
(3) Tree-LSTM \cite{tai2015improved}, which is a recursive model that generalizes the LSTM to arbitrary tree structures.
We investigate the child-sum variant of Tree-LSTM, and apply it to the dependency tree (or part of it).
In practice, we find that modifying this model by concatenating dependency label embeddings to the input of forget gates improves its performance on relation extraction, and therefore use this variant in our experiments.
Earlier, our group compared (1) and (2) with sequence models \cite{zhang2017tacred}, and we report these results; for (3) we report results with our own implementation.
\paragraph*{Neural sequence model.} Our group presented a competitive sequence model that employs a position-aware attention mechanism over LSTM outputs (PA-LSTM), and showed that it outperforms several CNN and dependency-based models by a substantial margin \cite{zhang2017tacred}.
We compare with this strong baseline, and use its open implementation in further analysis.%
\footnote{\href{https://github.com/yuhaozhang/tacred-relation}{ https://github.com/yuhaozhang/tacred-relation}}

\subsection{Experimental Setup}

We conduct experiments on two relation extraction datasets:
(1) \textbf{TACRED}:
Introduced in \cite{zhang2017tacred}, TACRED contains over 106k mention pairs drawn from the yearly TAC KBP\footnote{\href{https://tac.nist.gov/2017/KBP/index.html}{https://tac.nist.gov/2017/KBP/index.html}} challenge.
It represents 41 relation types and a special \emph{no\_relation} class when the mention pair does not have a relation between them within these categories.
Mentions in TACRED are typed, with subjects categorized into person and organization, and objects into 16 fine-grained types (e.g., date and location).
We report micro-averaged \fone{} scores on this dataset as is conventional.
(2) \textbf{SemEval 2010 Task 8}: The SemEval dataset is widely used in recent work, but is significantly smaller with 8,000 examples for training and 2,717 for testing.
It contains 19 relation classes over untyped mention pairs: 9 directed relations and a special \emph{Other} class.
On SemEval, we follow the convention and report the official macro-averaged \fone{} scores.

For fair comparisons on the TACRED dataset, we follow the evaluation protocol used in \cite{zhang2017tacred} by selecting the model with the median dev \fone{} from 5 independent runs and reporting its test \fone{}.
We also use the same ``entity mask'' strategy where we replace each subject (and object similarly) entity with a special \emph{SUBJ-$<$NER$>$} token.
For all models, we also adopt the ``multi-channel'' strategy by concatenating the input word embeddings with POS and NER embeddings.

Traditionally, evaluation on SemEval is conducted without entity mentions masked. However, as we will discuss in \refsec{semeval-bias}, this method encourages models to overfit to these mentions and fails to test their actual ability to generalize. We therefore report results with two evaluation protocols: (1) \emph{with-mention}, where mentions are kept for comparison with previous work; and (2) \emph{mask-mention}, where they are masked to test the generalization of our model in a more realistic setting.

Due to space limitations, we report model training details in the supplementary material.

\subsection{Results on the TACRED Dataset}

\begin{table}[!t]
	\centering
	\begin{tabular}{lccc}
		\toprule
		System & P & R & \fone \\
		\midrule
		LR$^\dagger$ (Zhang+\citeyear{zhang2017tacred}) & \fu{\fb{73.5}} & \f{49.9} & \f{59.4}\phantom{$^*$} \\
		SDP-LSTM$^\dagger$ (Xu+\citeyear{xu2015classifying}) & \f{66.3} & \f{52.7} & \f{58.7}\phantom{$^*$} \\
		Tree-LSTM$^\ddagger$ (Tai+\citeyear{tai2015improved}) & \f{66.0} & \f{59.2} & \f{62.4}\phantom{$^*$} \\
		PA-LSTM$^\dagger$ (Zhang+\citeyear{zhang2017tacred}) &\f{65.7} & \fu{64.5} & \f{65.1}\phantom{$^*$} \\
		\midrule
		GCN & \red{\f{69.8}} & \red{\f{59.0}} & \red{\f{64.0}}\phantom{$^*$}
		\\
		C-GCN & \red{\f{69.9}} & \red{\f{63.3}} & \red{\fu{66.4}}$^*$ \\ 
		\midrule
		GCN + PA-LSTM & \red{\f{71.7}} & \red{\f{63.0}} & \red{\f{67.1}}$^*$ \\ 
		C-GCN + PA-LSTM & \red{\f{71.3}} & \red{\fb{65.4}} & \red{\fb{68.2}}$^*$ \\ 
		\bottomrule
	\end{tabular}
\caption{Results on TACRED.
Underscore marks highest number among single models; bold marks highest among all.
$\dagger$ marks results reported in~\cite{zhang2017tacred}; $\ddagger$ marks results produced with our implementation.
$*$ marks statistically significant improvements over PA-LSTM with $p<.01$ under a bootstrap test.}
\label{tab:tacred}
\end{table}

We present our main results on the TACRED test set in \reftab{tacred}.
We observe that our GCN model outperforms all dependency-based models by at least 1.6 \fone.
By using contextualized word representations, the C-GCN model further outperforms the strong PA-LSTM model by 1.3 \fone, and achieves a new state of the art.
In addition, we find our model improves upon other dependency-based models in both precision and recall.
Comparing the C-GCN model with the GCN model, we find that the gain mainly comes from improved recall.
We hypothesize that this is because the C-GCN is more robust to parse errors by capturing local word patterns (see also \refsec{syntactic}).

As we will show in \refsec{syntactic}, we find that our GCN models have complementary strengths when compared to the PA-LSTM.
To leverage this result, we experiment with a simple interpolation strategy to combine these models.
Given the output probabilities $P_{G}(r | x)$ from a GCN model and $P_{S}(r | x)$ from the sequence model for any relation $r$, we calculate the interpolated probability as
\begin{align*}
P(r | x) &= \alpha \cdot P_{G}(r | x) + (1 - \alpha) \cdot P_{S}(r | x)
\end{align*}
where $\alpha \in [0, 1]$ is chosen on the dev set and set to 0.6.
This simple interpolation between a GCN and a PA-LSTM achieves an \fone{} score of 67.1,
outperforming each model alone by at least 2.0 \fone{}.
An interpolation between a C-GCN and a PA-LSTM further improves the result to 68.2.

\subsection{Results on the SemEval Dataset} \label{sec:semeval-results}

\begin{table}[!t]
	\centering
	\begin{tabular}{lcc}
		\toprule
		System & \textit{with-m} & \textit{mask-m} \\
		\midrule
		SVM$^\dagger$ (Rink+\citeyear{rink2010utd}) & \f{82.2}\phantom{$^*$} & --\phantom{$^*$} \\
		SDP-LSTM$^\dagger$ (Xu+\citeyear{xu2015classifying}) & \f{83.7}\phantom{$^*$} & --\phantom{$^*$} \\
		SPTree$^\dagger$ (Miwa+\citeyear{miwa2016end}) & \f{84.4}\phantom{$^*$} & --\phantom{$^*$} \\
		PA-LSTM$^\ddagger$ (Zhang+\citeyear{zhang2017tacred}) & \red{\f{82.7}}\phantom{$^*$} & \red{\f{75.3}}\phantom{$^*$} \\
		\midrule
		Our Model (C-GCN) & \red{\fb{84.8}}$^*$ & \red{\fb{76.5}}$^*$ \\
		\bottomrule
	\end{tabular}
	\caption{\fone{} scores on SemEval.
	$\dagger$ marks results reported in the original papers;
	$\ddagger$ marks results produced by using the open implementation.
	The last two columns show results from \emph{with-mention} evaluation and \emph{mask-mention} evaluation, respectively.
	$*$ marks statistically significant improvements over PA-LSTM with $p<.05$ under a bootstrap test.}
	\label{tab:semeval}
\end{table}

To study the generalizability of our proposed model, we also trained and evaluated our best C-GCN model on the SemEval test set (\reftab{semeval}).
We find that under the conventional \emph{with-entity} evaluation, our C-GCN model outperforms all existing dependency-based neural models on this separate dataset.
Notably, by properly incorporating off-path information, our model  outperforms the previous shortest dependency path-based model (SDP-LSTM).
Under the \emph{mask-entity} evaluation, our C-GCN model also outperforms PA-LSTM by a substantial margin, suggesting its generalizability even when entities are not seen.

\subsection{Effect of Path-centric Pruning} \label{sec:prune_exp}

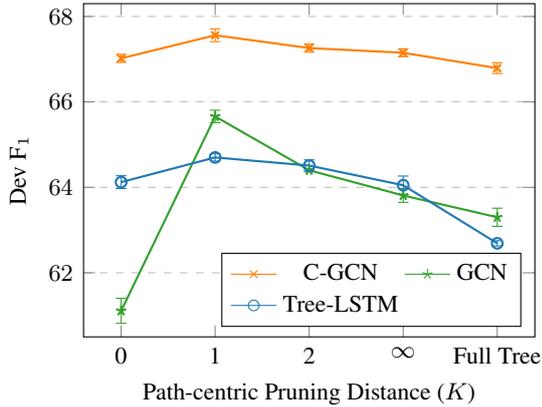
\begin{figure}
	\pgfplotstableread[row sep=\\,col sep=&]{
	x & f1 & std \\
0	&	64.122	&	0.150512458 \\
1	&	64.7	&	0.085029407 \\
2	&	64.51	&	0.13347659 \\
LCA	&	64.05	&	0.21488568542 \\
Full	&	62.69	&	0.0610 \\
}\treelstm

\pgfplotstableread[row sep=\\,col sep=&]{
	x & f1 & std \\
0	&	66.25	&	0.150576818 \\
1	&	66.52	&	0.093467641 \\
2	&	65.93	&	0.165647916 \\
LCA	&	66.13	&	0.230404444 \\
Full	&	65.83	&	0.113279204 \\
}\cgcn

\pgfplotstableread[row sep=\\,col sep=&]{
	x & f1 & std \\
0	&	67.02	&	0.09615612305 \\
1	&	67.56	&	0.1489295135 \\
2	&	67.26	&	0.09579144012 \\
LCA	&	67.15	&	0.09300537619 \\
Full	&	66.79	& 0.1253156016 \\
}\newcgcn

\pgfplotstableread[row sep=\\,col sep=&]{
		x & f1 & std \\
0	&	62.3	&	0.099281418 \\
1	&	65.14	&	0.071196404 \\
2	&	64.23	&	0.076071033 \\
LCA	&	63.61	&	0.031886329 \\
Full	&	63.2	&	0.079022642 \\
}\gcn

\pgfplotstableread[row sep=\\,col sep=&]{
		x & f1 & std \\
0	&	61.11	&	0.2938026549 \\
1	&	65.66	&	0.1467174155 \\
2	&	64.40	&	0.0339705755 \\
LCA	&	63.81	&	0.1608415369 \\
Full	&	63.30	&	0.2125417606 \\
}\newgcn

\begin{tikzpicture}[every plot/.append style={thick},font=\small]
\begin{axis}[
symbolic x coords={0, 1, 2, LCA, Full},
xticklabels={0, 0, 1, 2, $\infty$, Full Tree},
xlabel={Path-centric Pruning Distance ($K$)},
ylabel={Dev \fone},
ymin=60.5,
ymax=68.3,
height=6.0cm,
legend pos=south east,
legend columns=2,
ymajorgrids=true,
grid style=dashed,
]
\addplot[color=orange, mark=x] plot[error bars/.cd, y dir=both, y explicit] table[x=x,y=f1,y error=std]{\newcgcn};
\addplot[color=green, mark=star] plot[error bars/.cd, y dir=both, y explicit] table[x=x,y=f1,y error=std]{\newgcn};
\addplot[color=blue, mark=o] plot[error bars/.cd, y dir=both, y explicit] table[x=x,y=f1,y error=std]{\treelstm};
\legend{C-GCN, GCN, Tree-LSTM}

\end{axis}
\end{tikzpicture}
	\caption{Performance of dependency-based models under different pruning strategies. For each model we show the \fone{} score on the TACRED dev set averaged over 5 runs, and error bars indicate standard deviation of the mean estimate. $K=\infty$ is equivalent to using the subtree rooted at the LCA.} \label{fig:prune}
\end{figure}

To show the effectiveness of path-centric pruning, we compare the two GCN models and the Tree-LSTM when the pruning distance $K$ is varied.
We experimented with $K\in \{0,1,2,\infty\}$ on the TACRED dev set, and also include results when the full tree is used.
As shown in \reffig{prune}, the performance of all three models peaks when $K=1$, outperforming their respective dependency path-based counterpart ($K=0$).
This confirms our hypothesis in \refsec{prune} that incorporating off-path information is crucial to relation extraction.
\citet{miwa2016end} reported that a Tree-LSTM achieves similar performance when the dependency path and the LCA subtree are used respectively.
Our experiments confirm this, and further show that the result can be improved by path-centric pruning with $K = 1$.

We find that all three models are less effective when the entire dependency tree is present, indicating that including extra information hurts performance.
Finally, we note that contextualizing the GCN makes it less sensitive to changes in the tree structures provided, presumably because the model can use word sequence information in the LSTM layer to recover any off-path information that it needs for correct relation extraction.

\section{Analysis \& Discussion}
\label{sec:analysis}

\subsection{Ablation Study}
\label{sec:ablation}

To study the contribution of each component in the C-GCN model, we ran an ablation study on the TACRED dev set (\reftab{ablation}).
We find that:
(1) The entity representations and feedforward layers contribute 1.0 \fone{}.
(2) When we remove the dependency structure (i.e., setting $\tilde{\bA}$ to $\bI$), the score drops by 3.2 \fone{}.
(3) \fone{} drops by 10.3 when we remove the feedforward layers, the LSTM component and the dependency structure altogether.
(4) Removing the pruning (i.e., using full trees as input) further hurts the result by another 9.7 \fone{}.

\subsection{Complementary Strengths of GCNs and PA-LSTMs}
\label{sec:syntactic}

  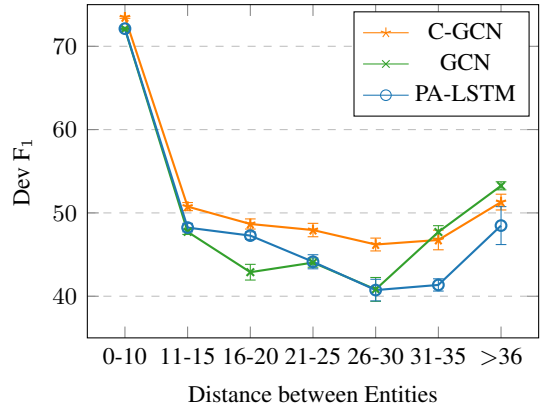
\begin{figure}[!t]
      \pgfplotstableread[row sep=\\,col sep=&]{
	x & f1 & std \\
	0	&72.1141326	&0.1744657 \\
	11	&48.2377529	&0.5087326 \\
	16	&47.2754772	&0.4516327 \\
	21	&44.1356432	&0.8444812 \\
	26	&40.7354675	&1.2974531 \\
	31	&41.3553217	&0.7352715 \\
	36	&48.4787961	&2.2773824 \\
}\rnnsurfdist

\pgfplotstableread[row sep=\\,col sep=&]{
		x & f1 & std \\
		0	&72.8164341	&0.1396928 \\
		11	&48.4548706	&0.5818213 \\
		16	&46.6699648	&0.4238466 \\
		21	&45.0233467	&0.8604094 \\
		26	&43.3434008	&1.0642209 \\
		31	&46.3724149	&0.9395862 \\
		36	&50.8828843	&1.3695015 \\
}\oldcgcnsurfdist
\pgfplotstableread[row sep=\\,col sep=&]{
		x & f1 & std \\
0	&73.09941158801692	&0.09608306093963896\\
11	&50.257997481938	&0.3452013459032025\\
16	&48.224387539099865	&0.9543582120727296\\
21	&48.03614128359533	&0.5039552440840683\\
26	&44.50581271114863	&1.2091378734481752\\
31	&47.80901829628172	&0.45116553750136623\\
36	&52.37074740638891	&1.5220435874138911\\
}\cgcnsurfdist

\pgfplotstableread[row sep=\\,col sep=&]{
		x & f1 & std \\
0	&73.47713584656135	&0.12246376688020903\\
11	&50.75629855271849	&0.48525771741020246\\
16	&48.67169674216086	&0.6068024788749192\\
21	&47.94355189046473	&0.8056728331902072\\
26	&46.20628929564897	&0.7698994395470727\\
31	&46.75173749770249	&1.1788646755148777\\
36	&51.30008544115039	&0.9465599383606671\\
}\newcgcnsurfdist

\pgfplotstableread[row sep=\\,col sep=&]{
		x & f1 & std \\
		0	&71.34742050328447	&0.06079241264245553\\
11	&47.71521942012715	&0.4211410634129581\\
16	&44.77980244124634	&0.5232909937881856\\
21	&43.801084324326446	&0.6912665784251945\\
26	&40.73786037440462	&0.6998341945504295\\
31	&45.43876916701202	&1.262304577940885\\
36	&51.686776949934846	&0.6983083505309853\\
}\gcnsurfdist

\pgfplotstableread[row sep=\\,col sep=&]{
		x & f1 & std \\
0	&72.07408605098267	&0.06912008871306403\\
11	&47.7288574871401	&0.30512796883835447\\
16	&42.88417908161794	&0.9429789338308927\\
21	&44.035453400223986	&0.5534139675767392\\
26	&40.81965839164089	&1.4299815732019252\\
31	&47.70695971295534	&0.7601092891376169\\
36	&53.25472728938262	&0.4846002014030102\\
}\newgcnsurfdist

\begin{tikzpicture}[every plot/.append style={thick},font=\small]
\begin{axis}[
symbolic x coords={0, 11, 16, 21, 26, 31, 36},
xtick=data,
ymin=35.0,
ymax=75.0,
height=6.0cm,
ytick = {40, 50, 60, 70},
xticklabels = {0-10, 11-15, 16-20, 21-25, 26-30, 31-35, $>$36},
xlabel={Distance between Entities},
ylabel={Dev \fone},
ymajorgrids=true,
grid style=dashed,
]
\addplot[color=orange, mark=star] plot[error bars/.cd, y dir=both, y explicit] table[x=x,y=f1,y error=std]{\newcgcnsurfdist};
\addplot[color=green, mark=x] plot[error bars/.cd, y dir=both, y explicit] table[x=x,y=f1,y error=std]{\newgcnsurfdist};
\addplot[color=blue, mark=o] plot[error bars/.cd, y dir=both, y explicit] table[x=x,y=f1,y error=std]{\rnnsurfdist};
\legend{C-GCN, GCN, PA-LSTM}

\end{axis}
\end{tikzpicture}
      \caption{Dev set performance with regard to distance between the entities in the sentence for C-GCN, GCN and PA-LSTM.
      Error bars indicate standard deviation of the mean estimate over 5 runs.
      } \label{fig:surfdist}
  \end{figure}

\begin{table}
    \centering
    \begin{tabular}{lc}
      \toprule
      Model   & Dev \fone \\
      \midrule
      \model{Best C-GCN} & \f{67.4}\\
          \model{\quad-- $h_{\text{s}}$, $h_{\text{o}}$, and Feedforward (FF)} & \f{66.4}\\
      \model{\quad-- LSTM Layer} & \f{65.5} \\
      \model{\quad-- Dependency tree structure} & \f{64.2} \\
          \model{\quad-- FF, LSTM, and Tree} & \f{57.1} \\
          \model{\quad-- FF, LSTM, Tree, and Pruning} & \f{47.4} \\
      \bottomrule
    \end{tabular}
    \caption{An ablation study of the best C-GCN model. Scores are median of 5 models.}
  \label{tab:ablation}
  \end{table}

  \begin{figure*}[!t]
      \centering
      \includegraphics[width=0.95\textwidth]{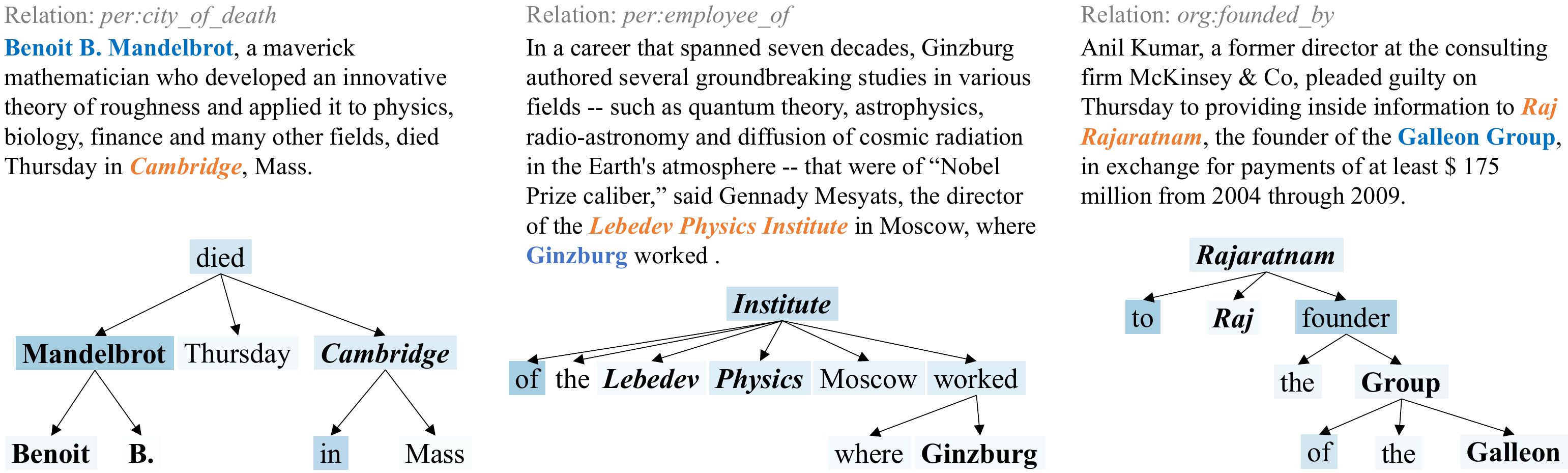}
      \vspace{-.3em}
      \caption{
      Examples and the pruned dependency trees where the C-GCN predicted correctly.
      Words are shaded by the number of dimensions they contributed to $h_{\text{sent}}$ in the pooling operation, with punctuation omitted.}
      \label{fig:casestudy}
  \end{figure*}

  \begin{table*}
      \centering
      \small
      \setlength{\tabcolsep}{12pt}
      \begin{tabular}{lccc}
          \toprule
          Relation & \multicolumn{3}{c}{Dependency Tree Edges}\\
          \midrule
          \rel{per:children} & \ner{S-PER} $\gets$ son & son $\to$ \ner{O-PER} &\ner{S-PER} $\gets$ survived \\
          \rel{per:other\_family} & \ner{S-PER} $\gets$ stepson & niece $\to$ \ner{O-PER} & \ner{O-PER} $\gets$ stepdaughter \\
          \rel{per:employee\_of} & a $\gets$ member & \ner{S-PER} $\gets$ worked & \ner{S-PER} $\gets$ played \\
          \rel{per:schools\_attended} & \ner{S-PER} $\gets$ graduated & \ner{S-PER} $\gets$ earned & \ner{S-PER} $\gets$ attended \\

          \rel{org:founded} & founded $\to$ \ner{O-DATE} & established $\to$ \ner{O-DATE} & was $\gets$ founded \\
          \rel{org:number\_of\_employees} & \ner{S-ORG} $\gets$ has & \ner{S-ORG} $\to$ employs & \ner{O-NUMBER} $\gets$ employees \\
          \rel{org:subsidiaries} & \ner{S-ORG} $\gets$ \ner{O-ORG} & \ner{S-ORG} $\to$ 's & \ner{O-ORG} $\to$ division \\
          \rel{org:shareholders} & buffett $\gets$ \ner{O-PER} & shareholder $\to$ \ner{S-ORG} & largest $\gets$ shareholder \\
          \bottomrule
      \end{tabular}
      \caption{The three dependency edges that contribute the most to the classification of different relations in the TACRED dev set.
	For clarity, we removed edges which 1) connect to common punctuation (i.e., commas, periods, and quotation marks), 2) connect to common prepositions (i.e., of, to, by), and 3) connect between tokens within the same entity.
	We use \ner{PER}, \ner{ORG} for entity types of \ner{PERSON}, \ner{ORGANIZATION}.
	We use \ner{S-} and \ner{O-} to denote subject and object entities, respectively.
	We also include edges for more relations in the supplementary material.
	}
      \label{tab:edges}
  \end{table*}

To understand what the GCN models are capturing and how they differ from a sequence model such as the PA-LSTM, we compared their performance over examples in the TACRED dev set.
Specifically, for each model, we trained it for 5 independent runs with different seeds, and for each example we evaluated the model's accuracy over these 5 runs.
For instance, if a model correctly classifies an example for 3 out of 5 times, it achieves an accuracy of 60\% on this example.
We observe that on 847 (3.7\%) dev examples, our C-GCN model achieves an accuracy at least 60\% higher than that of the PA-LSTM, while on 629 (2.8\%) examples the PA-LSTM achieves 60\% higher.
This complementary performance explains the gain we see in \reftab{tacred} when the two models are combined.

We further show that this difference is due to each model's competitive advantage (\reffig{surfdist}):
dependency-based models are better at handling sentences with entities farther apart, while sequence models can better leverage local word patterns regardless of parsing quality (see also \reffig{cgcn_vs_palstm}).
We include further analysis in the supplementary material.

\begin{sidewaysfigure}
    \centering
    \vspace{-4em}
    \input{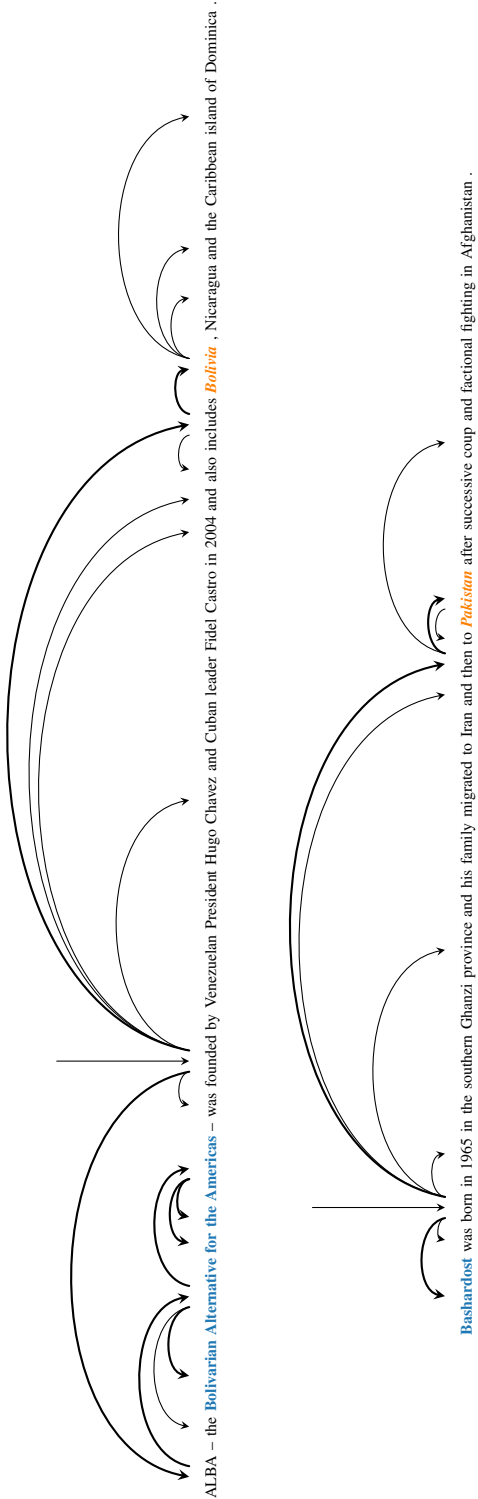}
    \caption{Dev set examples where either the C-GCN (upper) or the PA-LSTM (lower) predicted correctly in five independent runs. For each example, the predicted and pruned dependency tree corresponding to $K=1$ in path-centric pruning is shown, and the shortest dependency path is thickened. We omit edges to punctuation for clarity. The first example shows that the C-GCN is effective at leveraging long-range dependencies while reducing noise with the help of pruning (while the PA-LSTM predicts \textit{no\_relation} twice, \textit{org:alternate\_names} twice, and \textit{org:parents} once in this case). The second example shows that the PA-LSTM is better at leveraging the proximity of the word ``migrated'' regardless of attachment errors in the parse (while the C-GCN is misled to predict \textit{per:country\_of\_birth} three times, and \textit{no\_relation} twice).} \label{fig:cgcn_vs_palstm}
\end{sidewaysfigure}

\subsection{Understanding Model Behavior}
\label{sec:visualization}

To gain more insights into the C-GCN model's behavior, we visualized the partial dependency tree it is processing and how much each token's final representation contributed to $h_\text{sent}$ (\reffig{casestudy}).
We find that the model often focuses on the dependency path, but sometimes also incorporates off-path information to help reinforce its prediction.
The model also learns to ignore determiners (e.g., ``the'') as they rarely affect relation prediction.

To further understand what dependency edges contribute most to the classification of different relations, we scored each dependency edge by summing up the number of dimensions each of its connected nodes contributed to $h_\text{sent}$.
We present the top scoring edges in \reftab{edges}.
As can be seen in the table, most of these edges are associated with indicative nouns or verbs of each relation.\footnote{We do notice the effect of dataset bias as well: the name ``Buffett'' is too often associated with contexts where shareholder relations hold, and therefore ranks top in that relation.}

\subsection{Entity Bias in the SemEval Dataset}
\label{sec:semeval-bias}

In our study, we observed a high correlation between the entity mentions in a sentence and its relation label in the SemEval dataset.
We experimented with PA-LSTM models to analyze this phenomenon.%
\footnote{We choose the PA-LSTM model because it is more amenable to our experiments with simplified examples.}
We started by simplifying every sentence in the SemEval training and dev sets to ``\emph{subject} and \emph{object}'', where \emph{subject} and \emph{object} are the actual entities in the sentence.
Surprisingly, a trained PA-LSTM model on this data is able to achieve \red{65.1} \fone{} on the dev set if GloVe is used to initialize word vectors, and \red{47.9} dev \fone{} even without GloVe initialization.
To further evaluate the model in a more realistic setting, we trained one model with the original SemEval training set (unmasked) and one with mentions masked in the training set, following what we have done for TACRED (masked).
While the unmasked model achieves a \red{83.6} \fone{} on the original SemEval dev set, \fone{} drops drastically to \red{62.4} if we replace dev set entity mentions with a special \emph{$<$UNK$>$} token to simulate the presence of unseen entities.
In contrast, the masked model is unaffected by unseen entity mentions and achieves a stable dev \fone{} of \red{74.7}.
This suggests that models trained without entities masked generalize poorly to new examples with unseen entities.
Our findings call for more careful evaluation that takes dataset biases into account in future relation extraction studies.

\section{Conclusion}

We showed the success of a neural architecture based on a graph convolutional network for relation extraction.
We also proposed path-centric pruning to improve the robustness of dependency-based models by removing irrelevant content without ignoring crucial information.
We showed through detailed analysis that our model has complementary strengths to sequence models, and that the proposed pruning technique can be effectively applied to other dependency-based models.

\section*{Acknowledgements}

We thank Arun Chaganty, Kevin Clark, Sebastian Schuster, Ivan Titov, and the anonymous reviewers for their helpful suggestions. This material is based in part upon work supported by the National Science Foundation under Grant No.\ IIS-1514268. Any opinions, findings, and conclusions or recommendations expressed in this material are those of the authors and do not necessarily reflect the views of the National Science Foundation.

\bibliography{main}
\bibliographystyle{acl_natbib_nourl}

\clearpage
\appendix
\section{Experimental Details}
\label{sec:detail}

\subsection{Hyperparameters}
\paragraph*{TACRED}
We set LSTM hidden size to 200 in all neural models.
We also use hidden size 200 for the output feedforward layers in the GCN model.
We use 2 GCN layers and 2 feedforward (FFNN) layers in our experiments.
We employ the ReLU function for all nonlinearities in the GCN layers and the standard max pooling operations in all pooling layers.
For the Tree-LSTM model, we find a 2-layer architecture works substantially better than the vanilla 1-layer model, and use it in all our experiments.
For both the Tree-LSTM and our models, we apply path-centric pruning with $K = 1$, as we find that this generates best results for all models (also see \reffig{prune}).
We use the pre-trained 300-dimensional GloVe vectors~\cite{pennington2014glove} to initialize word embeddings, and we use embedding size of 30 for all other embeddings (i.e., POS, NER).
We use the dependency parse trees, POS and NER sequences as included in the original release of the dataset, which was generated with Stanford CoreNLP \cite{manning2014stanford}.
For regularization we apply dropout with $p=0.5$ to all LSTM layers and all but the last GCN layers.

\paragraph*{SemEval}
\red{We use LSTM hidden size of 100 and use 1 GCN layer for the SemEval dataset.}
We preprocess the dataset with Stanford CoreNLP to generate the dependency parse trees, POS and NER annotations.
All other hyperparameters are set to be the same.

For both datasets, we work with the Universal Dependencies v1 formalism \cite{nivre2016universal}.

\subsection{Training}

For training we use Stochastic Gradient Descent with an initial learning rate of $1.0$.
We use a cutoff of 5 for gradient clipping.
For GCN models, we train every model for 100 epochs on the TACRED dataset, and from epoch 5 we start to anneal the learning rate by a factor of 0.9 every time the \fone{} score on the dev set does not increase after an epoch.
For Tree-LSTM models we find 30 total epochs to be enough.
Due to the small size of the SemEval dataset, we train all models for 150 epochs, and use an initial learning rate of 0.5 with a decay rate of 0.95.

In our experiments we found that the output vector $h_\text{sent}$ tends to have large magnitude, and therefore adding the following regularization term to the cross entropy loss of each example improves the results:
\begin{align}
\ell_{\text{reg}} & = \beta \cdot \| h_{\text{sent}} \|^2.
\label{eqn:lreg}
\end{align}
Here, $\ell_{\text{reg}}$ functions as an $l_2$ regularization on the learned sentence representations.
$\beta$ controls the regularization strength and we set $\beta = 0.003$.
We empirically found this to be more effective than applying $l_2$ regularization on the convolutional weights.

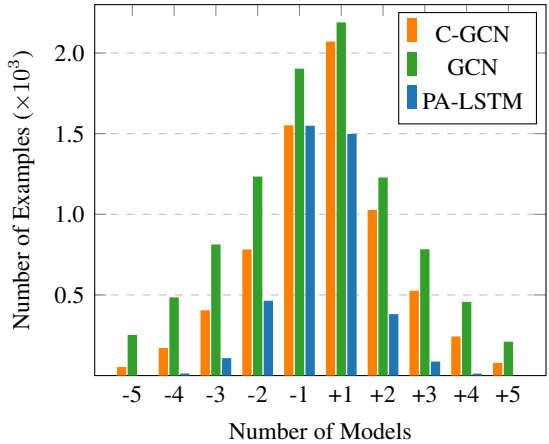
\begin{figure}
	\pgfplotstableread[row sep=\\,col sep=&]{
	x & cGCN & GCN & RNN \\
	-5 & 53	& 252	& 0 \\
	-4 & 171	& 485	& 13 \\
	-3 & 405	& 813	& 108 \\
	-2 & 782	& 1234	& 464 \\
	-1 & 1552	& 1903	& 1549 \\
	+1 & 2071	& 2190	& 1498 \\
	+2 & 1027	& 1228	& 381 \\
	+3 & 526	& 783	& 87 \\
	+4 & 242	& 457	& 12 \\
	+5 & 79	& 210	& 0 \\
}\mydata

\pgfplotsset{compat=1.11,
	/pgfplots/ybar legend/.style={
		/pgfplots/legend image code/.code={%
			\draw[##1,/tikz/.cd,bar width=3.5pt,yshift=-0.2em,bar shift=0pt]
			plot coordinates {(0cm,0.8em)};},
	},
}

\begin{tikzpicture}[font=\small]
\begin{axis}[
ybar=.5pt,
symbolic x coords={-5, -4, -3, -2, -1, +1, +2, +3, +4, +5},
xtick={-5, -4, -3, -2, -1, +1, +2, +3, +4, +5},
ymin=0, ymax=2300,
ytick={500, 1000, 1500, 2000},
yticklabels={0.5, 1.0 , 1.5, 2.0},
xlabel={Number of Models},
ylabel={Number of Examples ($\times 10^3$)},
height=6.5cm,
/pgf/bar width=3.5pt,
xtick align=inside,
ymajorgrids=true,
grid style=dashed,
]

\addplot[draw opacity=0, fill=orange] table[x=x,y=cGCN]{\mydata};
\addplot[draw opacity=0, fill=green] table[x=x,y=GCN]{\mydata};
\addplot[draw opacity=0, fill=blue] table[x=x,y=RNN]{\mydata};
\legend{C-GCN, GCN, PA-LSTM}

\end{axis}
\end{tikzpicture}
	\caption{Aggregated 5-run difference compared to PA-LSTM on the TACRED dev set. For each example, if $X$ out of 5 GCN models predicted its label correctly and $Y$ PA-LSTM models did, it is aggregated in the bar labeled $X-Y$. ``0'' is  omitted due to redundancy.} \label{fig:ensemble}
\end{figure}

\begin{figure*}[!h]
	\centering
	\includegraphics[width=0.95\textwidth]{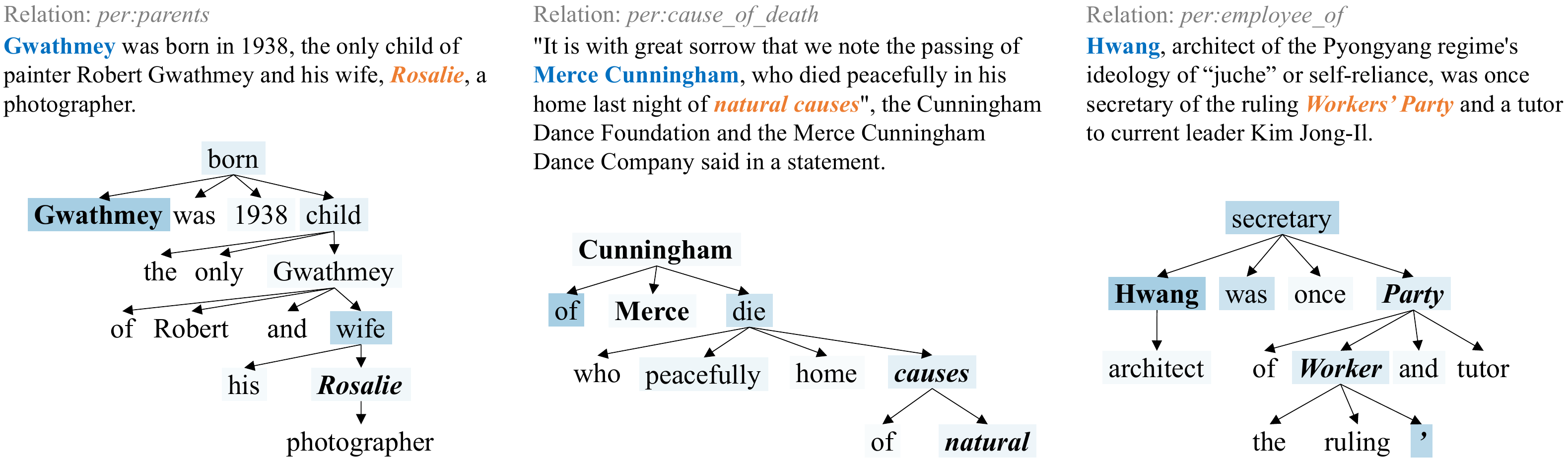}
	\caption{More examples and the pruned dependency trees the C-GCN predicted correctly.
	Words are shaded by the number of dimensions they contributed to $h_{\text{sent}}$ in the pooling operation, with punctuation omitted.}
	\label{fig:casestudy-more}
\end{figure*}

\begin{table*}[!h]
	\centering
	\small
	\begin{tabular}{lccc}
		\toprule
		Relation & \multicolumn{3}{c}{Dependency Tree Edges}\\
		\midrule
		\rel{per:children} & \ner{S-PER} $\gets$ son & son $\to$ \ner{O-PER} &\ner{S-PER} $\gets$ survived \\
		\rel{per:parents} & \ner{S-PER} $\gets$ born & \ner{O-PER} $\gets$ son & \ner{S-PER} $\gets$ mother \\
		\rel{per:siblings} & \ner{S-PER} $\gets$ sister & sister $\to$ \ner{O-PER} & brother $\to$ \ner{O-PER} \\
		\rel{per:other\_family} & \ner{S-PER} $\gets$ stepson & niece $\to$ \ner{O-PER} & \ner{O-PER} $\gets$ stepdaughter \\
		\rel{per:spouse} & wife $\to$ \ner{O-PER} & \ner{S-PER} $\gets$ wife & his $\gets$ wife \\
		\rel{per:city\_of\_death} & \ner{S-PER} $\gets$ died & died $\to$ \ner{O-CITY} & \ner{ROOT} $\to$ died \\
		\rel{per:city\_of\_birth} & \ner{S-PER} $\gets$ born & was $\gets$ born & born $\to$ \ner{O-CITY} \\
		\rel{per:cities\_of\_residence} & in $\gets$ \ner{O-CITY} & \ner{O-CITY} $\gets$ \ner{S-PER} & \ner{S-PER} $\gets$ lived \\
		\rel{per:employee\_of} & a $\gets$ member & \ner{S-PER} $\gets$ worked & \ner{S-PER} $\gets$ played \\
		\rel{per:schools\_attended} & \ner{S-PER} $\gets$ graduated & \ner{S-PER} $\gets$ earned & \ner{S-PER} $\gets$ attended \\
		\rel{per:title} & \ner{O-TITLE} $\gets$ \ner{S-PER} & as $\gets$ \ner{O-TITLE} & former $\gets$ \ner{S-PER} \\
		\rel{per:charges} & \ner{S-PER} $\gets$ charged & \ner{O-CHARGE} $\gets$ charges & \ner{S-PER} $\gets$ faces \\
		\rel{per:cause\_of\_death} & died $\to$ \ner{O-CAUSE} & \ner{S-PER} $\gets$ died & from $\gets$ \ner{O-CAUSE} \\
		\rel{per:age} & \ner{S-PER} $\to$ \ner{O-NUMBER} & \ner{S-PER} $\gets$ died & age $\to$ \ner{O-NUMBER} \\

		\rel{org:alternate\_names} & \ner{S-ORG} $\to$ \ner{O-ORG} & \ner{O-ORG} $\to$ $)$ & $($ $\gets$ \ner{O-ORG} \\
		\rel{org:founded} & founded $\to$ \ner{O-DATE} & established $\to$ \ner{O-DATE} & was $\gets$ founded \\
		\rel{org:founded\_by} & \ner{O-PER} $\to$ founder & \ner{S-ORG} $\gets$ \ner{O-PER} & founder $\to$ \ner{S-ORG} \\
		\rel{org:top\_members} & \ner{S-ORG} $\gets$ \ner{O-PER} & director $\to$ \ner{S-ORG} & \ner{O-PER} $\gets$ said \\
		\rel{org:subsidiaries} & \ner{S-ORG} $\gets$ \ner{O-ORG} & \ner{S-ORG} $\to$ 's & \ner{O-ORG} $\to$ division \\
		\rel{org:num\_of\_employees} & \ner{S-ORG} $\gets$ has & \ner{S-ORG} $\to$ employs & \ner{O-NUMBER} $\gets$ employees \\
		\rel{org:shareholders} & buffett $\gets$ \ner{O-PER} & shareholder $\to$ \ner{S-ORG} & largest $\gets$ shareholder \\
		\rel{org:website} & \ner{S-ORG} $\to$ \ner{O-URL} & \ner{ROOT} $\to$ \ner{S-ORG} & \ner{S-ORG} $\to$ :\\
		\rel{org:dissolved} & \ner{S-ORG} $\gets$ forced & forced $\to$ file & file $\to$ insolvency \\
		\rel{org:political/religious\_affiliation} & \ner{S-ORG} $\to$ group & \ner{O-IDEOLOGY} $\gets$ group & group $\to$ established \\
		\bottomrule
	\end{tabular}
	\caption{The three dependency edges that contribute the most to the classification of different relations in the dev set of TACRED. For clarity, we removed edges which 1) connect to common punctuation (i.e., commas, periods, and quotation marks), 2) connect to common preposition (i.e., of, to, by), and 3) connect tokens within the same entities. We use \ner{PER}, \ner{ORG}, \ner{CHARGE}, \ner{CAUSE} for entity types of \ner{PERSON}, \ner{ORGANIZATION}, \ner{CRIMINAL\_CHARGE} and \ner{CAUSE\_OF\_DEATH}, respectively. We use \ner{S-} and \ner{O-} to denote subject and object entities, respectively. \ner{ROOT} denotes the root node of the tree.
	}
	\label{tab:appendix-edges}
\end{table*}

\section{Comparing GCN models and PA-LSTM on TACRED} \label{sec:complementary}

We compared the performance of both GCN models with the PA-LSTM on the TACRED dev set.
To minimize randomness that is not inherent to these models, we accumulate statistics over 5 independent runs of each model, and report them in \reffig{ensemble}.
As is shown in the figure, both GCN models capture very different examples from the PA-LSTM model.
In the entire dev set of 22,631 examples, 1,450 had at least 3 more GCN models predicting the label correctly compared to the PA-LSTM, and 1,550 saw an improvement from using the PA-LSTM.
The C-GCN, on the other hand, outperformed the PA-LSTM by at least 3 models on a total of 847 examples, and lost by a margin of at least 3 on another 629 examples, as reported in the main text.
This smaller difference is also reflected in the diminished gain from ensembling with the PA-LSTM shown in \reftab{tacred}.
We hypothesize that the diminishing difference results from the LSTM contextualization layer, which incorporates more information readily available at the surface form, rendering the model's behavior more similar to a sequence model.

For reference, we also include in \reffig{ensemble} the comparison of another 5 different runs (with different seeds) of the PA-LSTM to the original 5 runs of the PA-LSTM.
This is to confirm that the difference shown in the figure between the model classes is indeed due a to model difference, rather than an effect of different random seeds.
More specifically, the two groups of PA-LSTM only see 99 and 121 examples exceeding the 3-model margin on either side over the 5 runs, much lower than the numbers reported above for the GCN models.

\section{Understanding Model Behavior} \label{sec:appendix-visual}

We present visualization of more TACRED dev set examples in \reffig{casestudy-more}.
We also show the dependency edges that contribute the most to more relation types in \reftab{appendix-edges}.

\end{document}